\documentclass[conference]{IEEEtran}

\usepackage{hyperref}
\usepackage{graphicx}
\usepackage{xcolor}
\usepackage{amsmath,amssymb,amsfonts,mathtools}
\usepackage{algorithmic}
\usepackage{bbm}
\usepackage[linesnumbered,ruled,vlined, french, onelanguage]{algorithm2e}
\SetKwInput{KwInput}{Entrée}     
\SetKwInput{KwOutput}{Sortie}     
\usepackage{multirow}
\usepackage{nccmath}
\usepackage{hyperref}
\usepackage{graphicx}
\usepackage{colortbl}
\usepackage{subfigure}
\usepackage{stfloats}

\usepackage{algorithmic}
\usepackage{multirow}
\usepackage{nccmath}
\begin{document}

\title{Viewing the process of generating counterfactuals as a source of knowledge: a new approach for explaining classifiers}


\author{\IEEEauthorblockN{Vincent Lemaire}
\IEEEauthorblockA{Orange Innovation Lannion, France\\
Email: vincent.lemaire@orange.com} \\
\IEEEauthorblockN{Victor Guyomard}
\IEEEauthorblockA{Orange Innovation, Lannion, France\\
Email: victor.guyomard@orange.com}
\and
\IEEEauthorblockN{Nathan Le Boudec}
\IEEEauthorblockA{Orange Innovation, Lannion, France\\ 
Email: nathan.leboudec@orange.com}\\ 
\IEEEauthorblockN{Françoise Fessant}
\IEEEauthorblockA{Orange Innovation, Lannion, France\\
Email: françoise.fessant@orange.com}}



\maketitle

\begin{abstract}
There are now many explainable AI methods for understanding the decisions of a machine learning model. Among these are those based on counterfactual reasoning, which involve simulating features changes and observing the impact on the prediction. 
This article proposes to view this simulation process as a source of creating a certain amount of knowledge that can be stored to be used, later, in different ways. This process is illustrated in the additive model and, more specifically, in the case of the naive Bayes classifier, whose interesting properties for this purpose are shown.
\end{abstract}


\section{Introduction}
\label{sec-introduction}

Machine learning, one of the branches of artificial intelligence, has enjoyed many successes in recent years. The decisions made by these models are increasingly accurate, but also increasingly more complex. However, it appears that some of these models are like black boxes: their decisions are difficult, if not impossible, to explain \cite{Bodria2023}. This lack of explainability can lead to a number of undesirable consequences: lack of user confidence, reduced usability of the models, presence of biases, etc. These needs have given rise to the field of XAI (eXplainable AI). XAI \cite{Saeed2021,allen2023interpretable}  is a branch of artificial intelligence that aims to make the decisions made by machine learning models intelligible, understandable, to users.

Among XAI methods, some of them are based on counterfactual reasoning. Counterfactual reasoning is a concept from psychology and social sciences \cite{xaisocial}, which involves examining possible alternatives to past events \cite{stepin2021survey}. Humans often use counterfactual reasoning by imagining what would happen if an event had not occurred, and this is precisely what counterfactual reasoning is. When applied to artificial intelligence, the question is, for example, ``Why did the model make this decision instead of another?" (counterfactual explanation) or ``How would the decision have differed if a certain condition had been changed?". This reasoning can take the form of a counterfactual or semi-factual explanation.

\begin{figure}[!ht]
    \centering
    \includegraphics[width=0.25\textwidth]{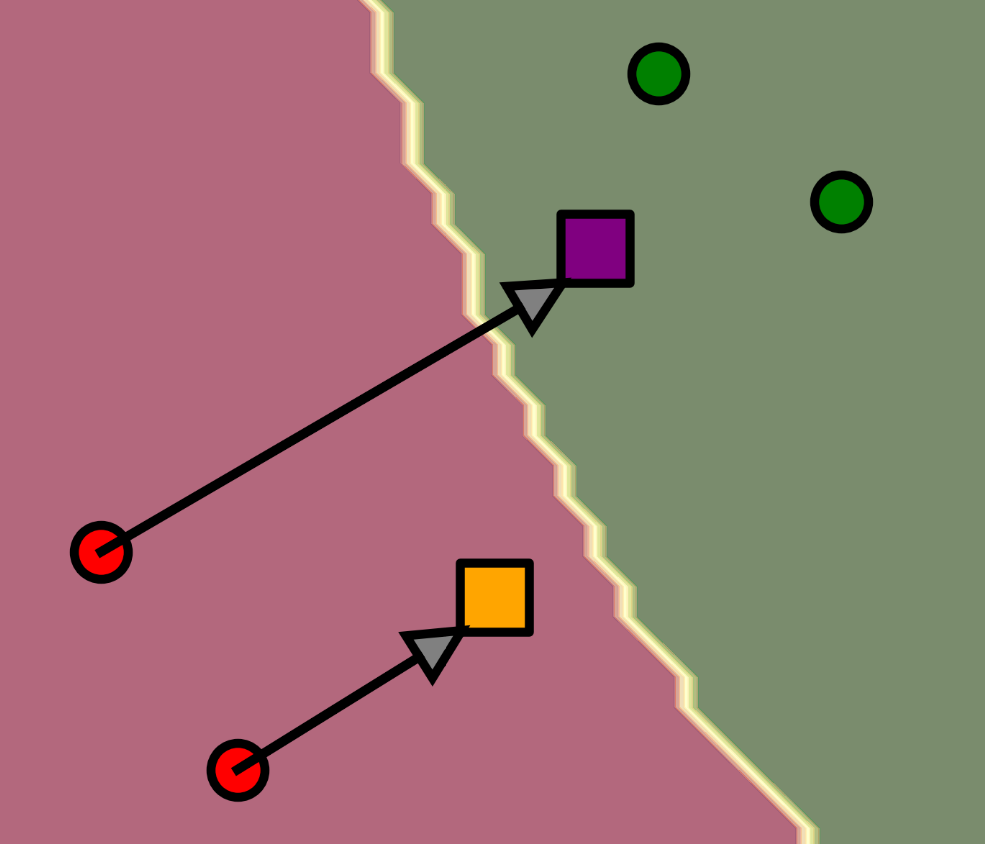}\
    \caption{Illustration of a counterfactual and a semi-factual. The red dots represent initial examples ($X$). The orange represents a semi-factual, the purple dot represents a counterfactual and the white line represents the decision boundary between the red and green classes.}
    \label{fig1}
\end{figure}

A counterfactual explanation might be ``If your income had been \$10000 higher, then your credit would have been accepted'' \cite{wachter2018counterfactual,LemaireIGI2010correlation,lemaire2009correlation}. A semi-factual is a special-case of the counterfactual in that it conveys possibilities that ``counteract'' what actually happened, even if the outcome does not change \cite{aryal2023explanations}: ``Even if your income had been \$5000 higher your credit would still be denied'' (but closer to being accepted), see Figure \ref{fig1}.

Within the framework of counterfactual reasoning, this article proposes to consider 
feature changes and their impacts on prediction as a source of knowledge that can be stored and exploited in various ways. This process is illustrated in the case of additive models and in particular in the case of the naive Bayes classifier, whose interesting properties for this purpose are shown.

The rest of this paper is organized as follows: Section \ref{sec-knowledge} explains (i) how to build a knowledge base from a classifier by using counterfactual reasoning and (ii) how to derive explanations from this knowledge base and (iii) some other additional usable knowledge. Section \ref{sec-nb} presents a concrete implementation for building the  knowledge base in the case of the naive Bayes classifier. The Section \ref{expe} illustrates, by means of an unsubscribing problem, (i) how clustering applied to this database generates new knowledge and (ii) examples of using the database to generate trajectories to transform the initial example to counterfactual or semi-factual. Finally the last section concludes the paper.

\section{Creation and use of a knowledge base}
\label{sec-knowledge}
In this section, we introduce the notations needed to understand the contribution. We then explain how to build a knowledge base from a classifier using counterfactual reasoning. Finally, we present the different types of explanations that can be derived from the knowledge base.

\subsection{Backgrounds}
Let $f$ be a predictive model that is trained using $N$ examples, each described by a set of $d$ explanatory variables (a vector $X=\{X_1, . ..., X_d\}$, derived from a $\mathcal{X}$ distribution) so as to predict a categorical target variable denoted $Y=\{y_1, ..., y_K\}$ derived from a $\mathcal{Y}$ distribution.

We assume that the dataset taken as input by the classifier is discretised, meaning that each variable $x_i$ falls into an interval of values $I_q$, where $q$ is the $q-th$ interval of values. 

\subsection{Creation of the knowledge base}
Our goal is to build a table that stores each variable change and its effect on prediction.
As the dataset is discretised, each variable change is represented by membership of a new value interval of values. To measure the difference in prediction, we propose to use the predicted probability returned by the classifier. Table \ref{Table-delta-1} gives an example of such a table in the case of two variables $X_1$ and $X_2$, discretised (or grouped) respectively into 2 and 3 intervals (groups) of values ($I$). 
For each individual $l$, on each row of the table, we store the differences in the values of the predicted probabilities, where $\Delta(X^{l}_{i,*\rightarrow m}, X^l)$ is the difference in the predicted probability for class 1 when the value of the variable $i$ goes from its initial value $`*'$ to the value of the interval (group) $m$ (simplified to $\Delta(X^{l}_{i,*\rightarrow m})$ in the table).

\begin{table*}[!ht]
\begin{center}
\begin{tabular}{|c|c|c|c|c|c|}
\hline
& \multicolumn{2}{|c|}{$X_1$}    &  \multicolumn{3}{|c|}{$X_2$} \\\hline
&  $I_1$  & $I_2$  & $I_1$ & $I_2$ & $I_3$ \\\hline
${X^1}$ & $\Delta({X^{1}_{1,*\rightarrow 1}})$ & $\Delta({X^{1}_{1,*\rightarrow 2}})$ & $\Delta({X^{1}_{2,*\rightarrow 1}})$ & $\Delta({X^{1}_{2,*\rightarrow 2}})$ & $\Delta({X^{1}_{2,*\rightarrow 3}}$ \\\hline
${X^2}$ & $\Delta({X^{2}_{2,*\rightarrow 1}})$ & $\Delta({X^{2}_{2,*\rightarrow 2}})$ & $\Delta({X^{2}_{2,*\rightarrow 1}})$ & $\Delta({X^{2}_{2,*\rightarrow 2}})$ & $\Delta({X^{2}_{2,*\rightarrow 3}})$ \\ \hline
\end{tabular}
\end{center}
\caption{Illustration of the knowledge base, here in the case of two variables and two examples.}
\label{Table-delta-1}
\end{table*}

To exploit this knowledge base effectively, the classifier used must have an additivity property with regard to prediction. This means that individual differences in predicted probabilities can be added together to obtain the overall difference in predicted probability. More formally, suppose $X_i$ and $X_j$ are two variables and $m,p$ are two new interval values for each variable, respectively.

Then a classifier is considered as additive if for each individual $l$:
$$\medmath{\Delta(X^{l}_{(i,*\rightarrow m)}, X^l) + \Delta(X^{l}_{(j,*\rightarrow p)}, X^l) = \Delta(X^{l}_{(i,*\rightarrow m)\cup(j,*\rightarrow p)}, X^l) } $$

where $\Delta(X^{l}_{(i,*\rightarrow m)\cup(j,*\rightarrow p)}, X^l)$ is the difference in predicted probability for class 1 when both variable $i$ and $j$ are changed. 

We will show in section \ref{sec-nb} that this property is valid for the naive Bayes classifier. 
However, this choice is not restrictive, since any other classifier with this property can be used.

\subsection{Deriving explanations from the knowledge base}
Once this knowledge base has been built up, different types of explanations can be derived, such as counterfactual or semi-factual explanations. 
In addition, we introduce two new types of explanation, namely trajectories and profile clustering.

\subsection{Counterfactual explanation}
In the context of explainable AI, a counterfactual explanation is defined as the smallest change in feature values that changes a model's prediction towards a given outcome. An example of such a counterfactual explanation might be \textit{``if your monthly income had been $500\$$ higher, your credit would have been accepted''}. The new example, obtained by changing the characteristic income by $500\$$, is called the ``counterfactual". 
Many methods have been proposed to generate counterfactuals, focusing on specific properties such as realism ~\cite{PawelczykWWW20,VanLooverenECML21,Guyomard2022VCNetAS}, actionability~\cite{ustun2019actionable,poyiadzi2020face}, sparsity ~\cite{brughmans2021nice,Wachter2017CounterfactualEW}, or robustness ~\cite{guyomard2023generating}.

\vspace{5mm}
\subsubsection{Generating sparse counterfactuals}
The sparsity property involves having counterfactuals with the smallest number of modified variables.  
Sparse explanations are interesting because they are more comprehensible for human \cite{miller2018explanation} and also more actionable, i.e. the change proposed by the counterfactual is easier to implement. 
Such explanations can easily be obtained by exploiting our knowledge base.
For a given individual, $X$, we simply read the value of the largest $\Delta$ and then, as we have the additivity property, read the second value of the largest $\Delta$ for a second variable and so on.
At each stage we check whether ($f(\mathbf{X'}) > 0.5$). If this is the case, the counterfactual has been found\footnote{We could, on the other hand, maximize the number of changes, but this is often of little practical interest}.

\vspace{5mm}
\subsubsection{Taking into account other properties }
In other cases, the objective may be to find the closest counterfactual, but under `business constraints' defined by the user. For example, the search for counterfactuals could be restricted to making changes only in adjacent intervals (e.g. intervals of close values). Given the table \ref{Table-delta-1}, we would be allowed to move people from interval 1 to 2 for the second variable, but not from 1 to 3. The user can also constrain the search by requiring that one of the variables must always be changed to a certain value, and so on. This type of constraint can easily be  taken into account and integrated into a counterfactual search algorithm using the proposed knowledge table.

The literature on counterfactuals sets out some interesting properties on the subject, such as (i) the notion of minimality: having a counterfactual that differs as little as possible from the original example; (ii) realism: the generated counterfactual must not contain changes that do not make sense from the point of view of the data (e.g. a decrease in the ``age" of an individual), also known as plausibility \cite{pmlr-v180-nemirovsky22a}; (iii) generation of counterfactuals that are similar to real examples or in dense regions of the class of interest, in order to have robust counterfactuals \cite{guyomard2023generating}.

It should be noted that a large proportion of these different properties can be easily obtained by exploiting the knowledge base in a different way, since the user can select the variables on which he wishes to intervene, according to the criterion of his choice.

\subsection{Additional usable knowledge}
\subsubsection{Preventive and reactive actions - }
\label{prev}

So far, we have mainly talked about creating counterfactuals to explain the model's decisions (as mentioned in the introduction to this article), but also potentially to be able to take reactive actions. For example, if a bank customer is predicted to ``leave" (churner), the counterfactual example indicates one or more actions to be taken to try to retain him: these actions are known as {\bf ``reactive'' actions}.

\begin{figure}[!ht]
    \centering
    \includegraphics[width=0.25\textwidth]{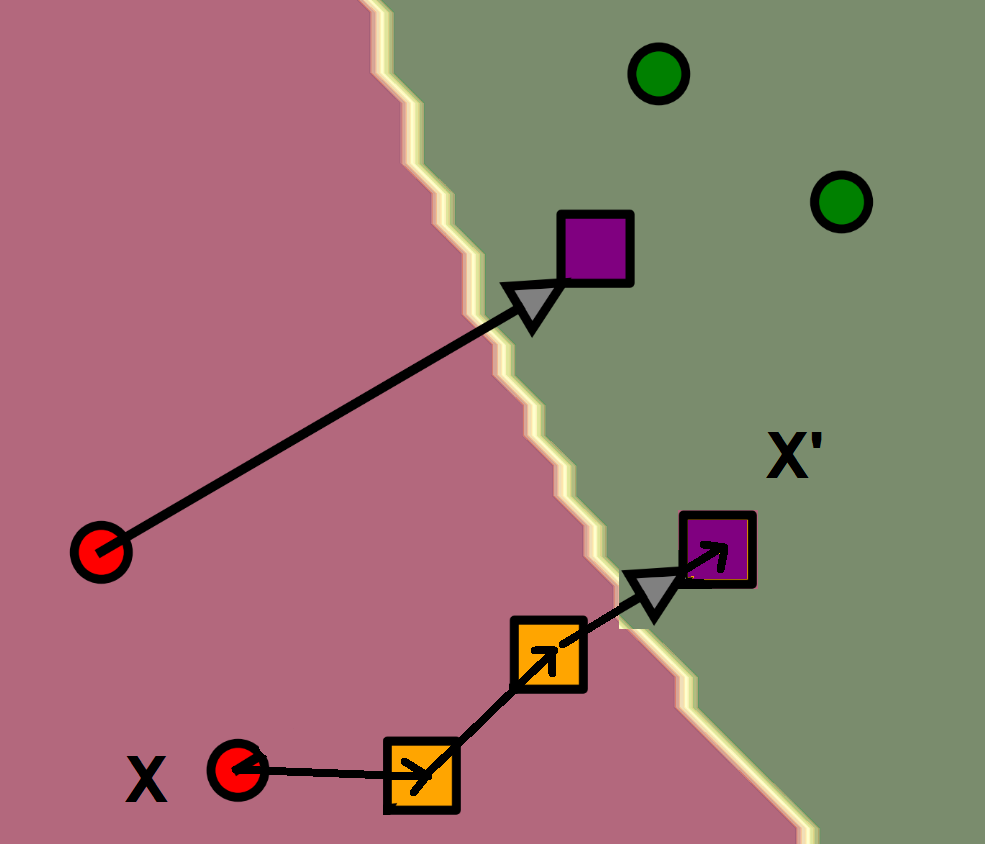}\
    \caption{Illustration of two counterfactuals: one achieved in 1 step, the second in 3 steps}
    \label{fig2}
\end{figure}

Conversely, the study of counterfactual trajectories a posteriori is of great interest, as it also makes it possible to identify when a trajectory is approaching the frontier (see Figure \ref{fig2}).

In such situations, reactive measures can be taken to reverse the trend and avoid undesirable outcomes (for example by observing the second step of the trajectory at the bottom of the figure \ref{fig2}). This approach is particularly relevant when it comes to predicting churn, for example, as it enables us to identify customers who are ``starting'' to churn. By being proactive, it is possible to implement targeted strategy to retain these customers and bring them back to a quality service.

\begin{figure}[!ht]
    \centering
    \includegraphics[width=0.25\textwidth]{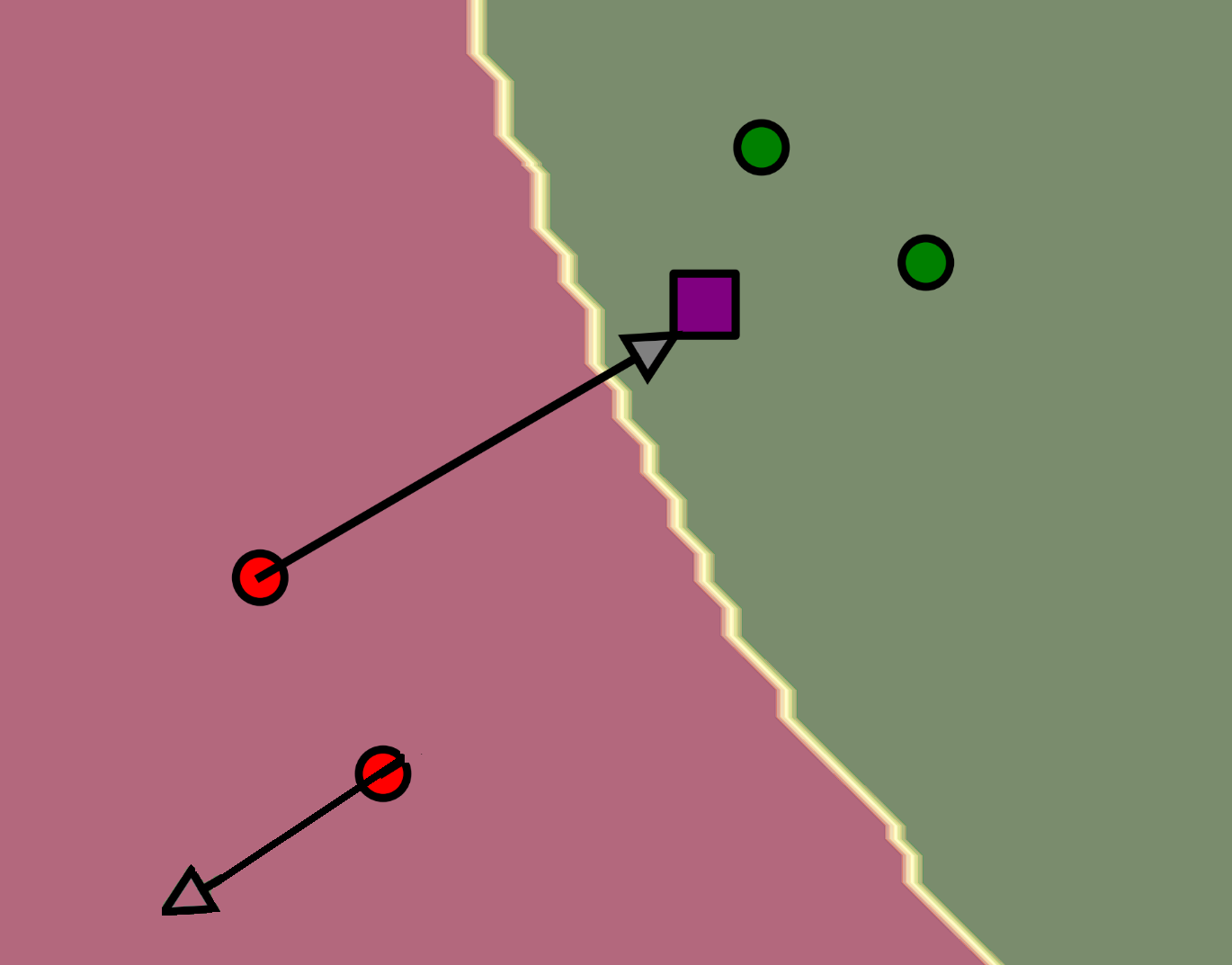}\
    \caption{Illustration of a counterfactuals (on top) and a ``negative'' semi-factual (bottom) which moves away from the decision frontier}
    \label{fig3}
\end{figure}

Finally, our knowledge base can also be used to carry out {\bf ``preventive" actions}. Going back to Figure \ref{fig1}, 
we can try to create a (negative) semi-factual which moves away from the decision frontier (see Figure \ref{fig3}) : "The customer is is not supposed to leave, but he is nevertheless close to the decision frontier". In this case, all we need to do is look at the negative values of $\Delta$ and move away from them  according to the user's wishes. For example, we could easily identify all the people who are one step away from crossing the decision boundary and act accordingly.

\subsubsection{Profile creation - }
The last way of using the knowledge base that we will describe here\footnote{The reader can imagine others: descriptive statistics of the table, number of individuals at 1, 2, 3 ... steps from the decision frontier, visualization of the trajectories, ... .}, is to carry out an exploratory analysis using a clustering technique.

Using the knowledge base, it is possible to group individuals according to the impact of each possible change, i.e. the impact resulting from each of these changes. Analysis of the clusters created can be a source of learning. This is illustrated in the next section. 

Note: all the uses of the knowledge base, presented in this paper, are on-demand, local, global or in-between,  so that both counterfactual and semi-counterfactual uses can be generated, allowing a wide range of uses.

\section{Computing $\Delta$ in the case of the naive Bayes classifier}
\label{sec-nb}

\subsection{Reminders on the naive Bayes classifier}

The naive Bayes classifier (NB) is a widely-used tool in supervised classification problems. It has the advantage of being efficient for many real data sets \cite{Hand2001}. However, the naive assumption of conditional independence of the variables  can, in some cases, degrade the classifier's performance. This is why variable selection methods have been developed \cite{Langley1994}. They mainly consist of variable addition and deletion heuristics to select the best subset of variables maximizing a classifier performance criterion, using a wrapper-type approach \cite{Guyon2003}. It has been shown in \cite{BoulleJMLR07} 
that averaging a large number of selective naive Bayes classifiers, performed with different subsets of variables, amounts to considering only one model with a weighting on the variables. Bayes' formula under the assumption of independence of the input variables conditionally to the class variable becomes:

\begin{equation}
P(C_{k}|X)=\frac{P(C_{k})\prod_{i}P(X_{i}|C_{k})^{W_i}}{\sum_{j=1}^{K}(P(C_{j})\prod_{i}P(X_{i}|C_{j})^{W_i})}
\label{eq-snb}
\end{equation}

where $W_i \in [0, 1]$ is the weight of variable $i$.
The predicted class is the one that maximizes the conditional probability $P(C_{k}|X)$. The probabilities $P(X_i|C_i)$ can be estimated by intervals using discretization for numerical variables. Gaussian naive Bayes could be also considered. For categorical variables, this estimation can be done directly if the variable takes few different modalities, or after grouping (of values) in the opposite case.

\subsection{Criteria to be optimized in the search for counterfactuals}
Let, $X$ as an example, and two classes $C_1$ and $C_2$. The search for a counterfactual consists in optimizing and increasing the probability of belonging to the target class $C_1$ when $X$ is initially predicted by the model to belong to $C_2$ (and vice versa). To do this, we can develop a gluttonous algorithm, which is expensive in terms of computation time and does not necessarily have the additivity properties described above.
We propose below to pose the problem differently, rewriting the equation \ref{eq-snb} and looking at how to increase the probability of belonging to a particular class of interest. To achieve this goal, and to maximize $P(C_j|X')$ with respect to the initial value of $P(C_j|X)$, we will exploit the following proposition:

If we take $X$ and $X'$ as two elements of the input space $\mathcal{X}$, we show that for a two-class classification problem, searching for counterfactuals of $X$ amounts to examining the evolution of the value of $\Delta$ when we change some of the values of $X$ to $X'$, such that:

\begin{equation}
\begin{split}
\Delta(X, X') &= \left( \sum_{i=1}^d W_i(\log{P(X_i|C_{1})}  - \log{P(X_i |C_{2})}) \right)\\ & - \left(  \sum_{i=1}^d W_i(\log{P(X'_i|C_{1})} - \log{P(X'_i |C_{2})}) \right)
\end{split}
\label{eq-delta}
\end{equation}

{\it Proof:} If we start again from the equation \ref{eq-snb}
\begin{equation}
P(C_j|{X}) = \frac{P(C_j)\prod_{i=1}^dP(X_i|C_j)^{W_j}}{\sum_{z}[P(C_z)\prod_{i=1}^dP(X_i|C_z)^{W_j}]}
\end{equation}
by posing: 

\begin{equation}
\begin{split}
L_j({X}) &=  \log \left( P(C_j)\prod_{i=1}^dP(X_i|C_j)^{W_i} \right) \\&= \log{P(C_j)} + \sum_{i=1}^d W_i\log{P(X_i|C_j)}
\end{split}
\end{equation}

then we have:
\begin{equation}
\begin{split}
P(C_j|{X}) &= \frac{e^{L_j({X})}}{\sum_{z}e^{L_z({X})}} \\&= \frac{1}{\sum_{z}e^{L_z({X}) - L_j({X})}} \\&= \frac{1}{1 + \sum_{z \neq j}e^{L_z({X}) - L_j({X})}}, 
\end{split}
\end{equation}

and so in the case of two classes:

\begin{equation}
P(C_j|{X}) = \frac{1}{1 + e^{L_{j'}({X}) - L_j({X})}}
\label{eq-norm}
\end{equation}

We can see that to get closer to the class $C_j$, all we have to do is reduce the quantity $L_{j'}({X}) - L_j({X})$, and thus reduce:
\begin{equation}
\begin{split}
\log{P(C_{j'})} + \sum_{i=1}^d W_i\log{P(X_i|C_{j'})} \\- \log{P(C_j)} - \sum_{i=1}^d W_i\log{P(X_i|C_j)}
\end{split}
\end{equation}

Since $P(C_{j})$ and $P(C_{j'})$ are constant, this is equivalent to decreasing: $$\sum_{i=1}^d W_i\log{P(X_i|C_{j'})} - \sum_{i=1}^d W_i\log{P(X_i|C_{j})}$$

and therefore to take an interest in the distance:

\begin{equation}
\begin{split}
\Delta({X}, {X'}) &= \sum_{i=1}^d W_i(\log{P(X_i  |C_{j'})} - \log{P(X_i |C_{j})}) \\
&- \sum_{i=1}^d W_i(\log{P(X'_i  |C_{j'})} - \log{P(X'_i  |C_{j})})
\end{split}
\label{delta_8}
\end{equation}

If $\Delta$ is positive then we are getting closer to the decision frontier (or even crossing it). If $\Delta$ is negative then we are moving away from the decision frontier and therefore away from the desired objective. The counterfactual search algorithm becomes straightforward. Simply calculate, for a given example $X$, the value of $\Delta$ for each explanatory variable and for each value of this explanatory variable. Then, given these values, iterate the successive changes in order to obtain a counterfactual example. These variable-by-variable changes have the property of being additive.  

Indeed, if we consider four examples $X^0$, $X^{'1}$, $X^{'2}$ and $X^{'3} \in \mathcal{X}$, which are respectively (i) an initial example $X^0$, then the same example for which we have modified only one explanatory variable $l$ for $X^{'1}$, $m$ for $X^{'2}$, and finally an example that cumulates the two univariate modifications $l$ and $m$ for $X^{'3}$, such that :

$$\exists! \ l \ \text{such as} \ X^{'1}_l \neq X^{0}_l$$
$$\exists! \ m \ \text{such as} \ X^{'2}_m \neq X^{0}_m \ \text{and} \ m \neq l$$
and $$ X^{'3}_k = \begin{cases}
    X^{'1}_l, & \text{if } k=l \\
    X^{'2}_m, & \text{if } k=m \\
    X^{0}_k, & \text{otherwise}
\end{cases}
$$

then it is obvious, from the additivity over all the variables in the equation \ref{eq-delta}, that we have : $\Delta({X_0},{X'_3}) = \Delta({X_0},{X'_1}) + \Delta({X_0},{X'_2})$. Modifying one variable and then the other is equivalent to modifying them simultaneously in the calculation of $\Delta$. It should also be noted that this additivity is demonstrated from the equation \ref{eq-norm}, so we can be sure of increasing the {\bf normalized} value of the probability of the class of interest, $P(C|X)$, which is a plus.

Note: the list of $\Delta$ values can potentially be very large if the number of distinct values of the explanatory variables is large. Nevertheless, it is common for some naive Bayes classifiers (except the Gaussian version) \cite{Yang2003,Yang2009} to discretize the numerical variables and group the modalities of the categorical variables, in a supervised manner, in order to obtain an estimate of the conditional densities ($P(X_i|C)$) which are then used in the calculation of $P(C|X)$. This is what has been done in this article using the \cite{BoulleML06} and \cite{BoulleJMLR05} methods for numerical and categorical variables respectively. These supervised discretization and grouping operations often produce a limited number of intervals or groups of modalities. This makes it possible to obtain a reasonable number of values to test.

\section{Illustration on an unsubscribe case}
\label{expe}

\subsection{Dataset and classifier used} 
\label{data}

This section uses the ``Telco Customer Churn" dataset provided by a fictitious telecommunications company that provided home telephone and internet services to 7043 customers in California. The aim is to classify people who may or may not leave the company. Each customer is described by 20 descriptive variables (3 numerical and 17 categorical) plus the class variable `churn' with two modalities (yes/no) with an unbalanced distribution (75\% non-churn). This dataset can be downloaded from Kaagle \cite{dataset-telco}.
We use 80\% of the data for learning and 20\% for testing. The naive Bayes classifier is produced using the Khiops library, which is available on Github \cite{khiops-web}, the rest of the computation is straightforward using Equation \ref{delta_8}.  

During the learning process, only 10 of the variables were retained in the model. Below are all the intervals of values or groups of modalities obtained during the pre-processing process (the value in brackets gives the weight of the variable in the model, equation \ref{snb}, values from 0 to 1):

\begin{itemize}
    \item 1 * - Tenure ($W_1$=0.67): {\small [0-0.5], ]0.5-1.5], ]1.5-5.5], ]5.5-17.5], ]17.5-42.5], ]42.5-58.5], ]58.5-71.5], ]71.5-72]}
    \item 2 - InternetService ($W_2$=0.78): {\small [Fiberoptic], [DSL], [No]}
    \item 3 - Contract ($W_3$=0.37): {\small[Month-to-month], [Twoyear], [Oneyear]}
    \item 4 - PaymentMethod ($W_4$=0.29): {\small [Mailedcheck], [Creditcard(automatic), Electroniccheck, Bank\-trans\-fer\-(au\-tomatic)]}
    \item 5 - OnlineSecurity ($W_5$=0.15): {\small [No], [Yes], [No internet service]}
    \item 6 -TotalCharges ($W_6$=0.29): {\small [18.8;69.225], ]69.225;91.2], ]91.2;347.9], ]347.9;1182.8], 
    
    ]1182.8 ;2065.7], ]2065.7;3086.8], ]3086.8;7859], ]7859;8684.8]}
    \item 7 * - PaperlessBilling ($W_7$=0.40): {\small [Yes], [No]}
    \item 8 - TechSupport ($W_8$=0.04): {\small [No], [Yes], [No internet service]}
    \item 9 * - SeniorCitizen ($W_9=0.28$): {\small [0], [1]}
    \item 10 *  - Dependents ($W_9=0.10$): {\small [Yes], [No]}
\end{itemize}

For all 10 variables, there are a total of 36 intervals/groupings and therefore 26  values of $\Delta$ to calculate in our knowledge base. Indeed, for each individual and each variable, there is a $\Delta$ value that has a null value, the value which corresponds to it factually and which therefore does not need to be calculated.

\subsection{Classifier analysis stage} 
\label{stage}

Before carrying out the clustering stage, it is important to take an interest in the variables retained during the classification model training stage. For example, although it may be interesting to analyze the `Tenure' variable, it is clearly not an actionable variable. Indeed, it is not possible to change a customer's seniority in order to make him potentially less unfaithful. The same applies to the `SeniorCitizen' and `Dependents' variables. We have also removed the `PaperlessBilling' variable, which has very little impact on the clustering results described below.
As a result, these 4 variables are not retained during the clustering stage below; only the informative, influential and actionable variables are retained \footnote{All the variables could have been retained but the clustering would have been biased by uninteresting variables from the point of view of creating counterfactual examples} (see Section \ref{prev}).

\subsection{Exploratory Analysis using a clustering}


\subsubsection{Clustering performed} 

The clustering performed is usual: (i) we use the table of $\Delta$ values calculated on the test set, (ii) we learn a k-means with the L2 \cite{Hartigan1979} distance for different $k$ values ($k \in \{2,12\}$), (iii) and finally we retain the k-means whose value of $k$ corresponds to the `elbow point', here $k=4$, \cite{Thorndike1953WhoBI} of the curve representing the global reconstruction distance versus the value of $k$. 

\begin{figure*}[!ht]
\centering
\includegraphics[width=1.0\textwidth]{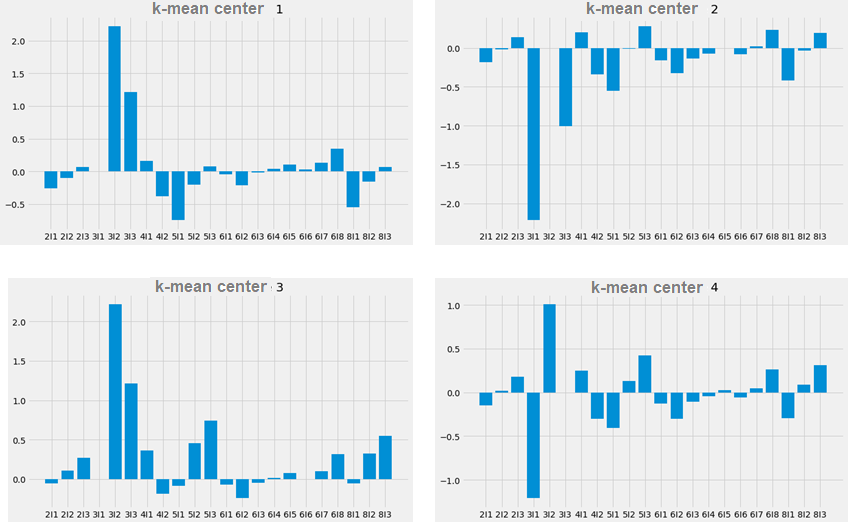}
\label{fig:cluster1centroid1}
\caption{Average profile of individuals in clusters represented as histograms. The names of the values on the abscissa refer to the number of the variables and the number of the intervals (groups) described above. For example, `3I2' refers to the third variable ('Contract') and its second interval / group of values ('Twoyear'). The ordinate values are the mean values of the cluster ($\Delta$).}
\label{4clusters}
\end{figure*}

\vspace{5mm}
\subsubsection{Results}

The resulting clusters are shown in Figure \ref{4clusters}. An analysis of these 4 clusters, combined with the predictions of the classifier, shows that:
\begin{itemize}
\item Cluster 1 (10\% of the global population and containing 2\% of customers predicted to be churner): individuals who can be made less churner mainly by means of variable 3 (`Contract') - i.e. by trying to get them to take out an annual contract (`Twoyear' or `OneYear'); NB - this marketing action is fairly difficult to carry out.
\item  Cluster 2 (24\% of the global population and containing no customer predicted to be churner): people who are very insensitive to the fact that they are becoming less churner (mostly negative $\Delta$ means). They may not be targeted by a `reactive' marketing campaign (which is in line with the classifier's predictions), but rather by a preventive campaign using the `contract' variable or the `payment method' variable (payment by card or direct debit).
\item Cluster 3 (45\% of the global population and containing 47\% of customers predicted to be churner (almost all of the individuals predicted to be churner)): some similarities with the individuals in cluster 1 for the `Contract' variable. On the other hand, we can see that the 5th ('OnlineSecurity') and 8th ('TechSupport') variables have a `leverage effect' in reducing churn. Offering a security or support option is very attractive to these individuals.
\item Cluster 4 (21\% of the population and containing no customers predicted to be unfaithful): individuals who are partially opposite to those in the first cluster, for example for the `Contract' variable, who should not be offered a `two-year contract' in this case.

\end{itemize}

The analysis of the clusters obtained here is not exhaustive. Indeed, it is an exploratory analysis where the data scientist and the business expert will spend the time needed to refine their joint analyses. However, the analysis carried out here allows us to identify interesting `reactive' actions to be taken with individuals in cluster 3 or preventive actions with individuals in cluster 2.

\subsection{Examples of trajectories}

For this Telco problem, a trajectory could be a customer going from ``no churn” to ``churn”, reciprocally from ``churn" to ``no churn", or staying no churner or churner. In this way, by understanding the trajectories approaching churn (or no churn), it would be possible to take preventive (reactive) action (see Section \ref{prev}) to slow down the trend towards churn (or to ``catch up'' the customer). 

The trajectories are presented as tables where the first line presents the initial profile (values of his input variables) of the customer. Then the following lines present each univariate change (cell in yellow), each step, in this trajectory. The number of lines differs from a customer to another one since some variables may not influence the $\Delta$ value and are therefore not included in the table. The columns in the table are the 6 variables kept (over the 10 included in the model) in the analysis (see Section \ref{stage}). 

We give here two representative trajectories extracted from our `knowledge base', one from ``no churn” toward ``churn” and one from "churn" to ``no churn":

\begin{itemize}
\item {\bf Trajectory 1 - ``no churn” toward ``churn” : } - In this case a customer moves from a ``no churn” situation to another ``no churn” situation but where he could be closer to the border (see Table \ref{table-1}). For this customer, 4 over the 6 variables allow to `walk' toward the border. At the end of the trajectory the customer remains 'loyal' (see the last column which indicate the probability to be a churner). The `customer' in the last line of this table is therefore the semi-factual of the original one. Detecting his movements towards the border can lead to preventive action. 
\vspace{6mm}

\begin{table*}[!]
\centering
\fontsize{9}{8}\selectfont
\begin{tabular}{|c|c|c|c|c|c|c|}
\hline
InternetSer	& OnlineSecu	& TechSup	& Contract	& PaymentM	& TotalCharges	 & Prob 'Yes' \\ \hline
DSL	& Yes	& No	& One year	        & M-c	& 1889.5		& 0.07 \\ \hline
DSL	& Yes	& No	& \cellcolor{yellow} M-t-M & M-c	& 1889.5		& 0.14 \\ \hline
DSL	& \cellcolor{yellow} No	& No	& \cellcolor{yellow} M-t-M	& M-c	& 1889.5		 & 0.17 \\ \hline
DSL	& \cellcolor{yellow} No	& No	& \cellcolor{yellow} M-t-M	& M-c	& \cellcolor{yellow} 80.275		& 0.28 \\ \hline
\cellcolor{yellow} Fiber optic	& \cellcolor{yellow} No	& No	& \cellcolor{yellow} M-t-M	& M-c	& \cellcolor{yellow} 80.275	& 0.48 \\ \hline
\end{tabular}
\caption{Illustration of a semi-factual - Trajectory 1 - ``no churn” toward ``churn” (Abbreviations used for column values (for place considerations) : 'M-t-M' for 'Month-to-month' , 'M-c' for 'Mailed check'), Abbreviations used for column names: InternetSer=InternetService, OnlineSecu=OnlineSecurity, TechSup=TechSupport, PaymentM=PaymentMethod}
\label{table-1}
\end{table*}

\item {\bf Trajectory 2 - ``churn” to ``no churn” : } - In this case a customer moves from a ``churn” situation to a ``no churn” situation; where he could cross the border (see Table \ref{table-2}). For this customer, 2 over the 6 variables allow  him to become a no churner. At the end of the trajectory the customer becomes 'loyal' (see the last column which indicate the probability to be a churner). One see that for him the variable which has the biggest impact is the 'PaymentMethod' while the one with the lowest impact is the 'TotalCharges'. It could be surprising that, for this customer, a bigger value of TotalCharges results in a lower probability to churn but an analysis of this variable confirms this interaction. Indeed the probability to churn when TotalCharges=80.275 (value belonging to the interval [18.8;69.225]) is larger than the one when TotalCharges=5036.3 (value belonging to the interval ]3086.8;7859]) after the preprocessing (see Section \ref{data}) used by the classifier\footnote{We do not give here all the statistics of the dataset but the reader may compute them easily by himself}. The `customer' in the last line of this table is a counterfactual of the original one. The yellow values are pieces of information to realize reactive actions.
\end{itemize}

Note: a future work will be to check whether the examples in the rows of the Tables \ref{table-1},\ref{table-2}, which are semifactuals or counterfactuals, belong to the density of the training examples; i.e their $P(X')$ is not outside (or too lower than) the values, $P(X)$, of the ones of factual examples. In the case of Naïve Bayes Classifier $P(X)$ could be simply be computed by $P(X)={\sum_{j=1}^{K}(P(C_{j})\prod_{i}P(X_{i}|C_{j})^{W_i})}
\label{snb} $ (the denominator of the equation \ref{eq-snb}).

\begin{table*}[!t]
\centering
\fontsize{9}{8}\selectfont
\begin{tabular}{|c|c|c|c|c|c|c|}
\hline
InternetSer	& OnlineSecu	& TechSup	& Contract	& PaymentM	& TotalCharges	 & Prob 'Yes' \\ \hline
Fiber optic	& No	& No	& M-t-M	& E-c	& 80.275	& 0.59 \\ \hline
Fiber optic	& No	& No	& M-t-M	& \cellcolor{yellow}B-t	& 80.275	& 0.50 \\ \hline
Fiber optic	& No	& No	& M-t-M& \cellcolor{yellow}B-t & \cellcolor{yellow}5036.3	&  0.33 \\ \hline
\end{tabular}
\caption{Illustration of a counterfactaul - Trajectory 2 - `` churn” to `` no churn” (Abbreviations used for column values (for place considerations) : 'M-t-M' for 'Month-to-month' , 'B-t' for 'Bank transfer', E-c for Electronic check), Abbreviations used for column names: InternetSer=InternetService, OnlineSecu=OnlineSecurity, TechSup=TechSupport, PaymentM=PaymentMethod}
\label{table-2}
\end{table*}

\section{Conclusion}
In the context of methods for explaining the results of a machine learning model, this article has proposed to consider the process of generating counterfactual examples as a source of knowledge that can be stored and then exploited in different ways. This process has been illustrated in the case of additive models and in particular in the case of the naive Bayes classifier, whose interesting properties for this purpose have been shown.
We have also suggested the quantities that can be stored and the different ways of exploiting them. Some of the results have been illustrated on a churn problem, but the approach is equally exploitable in other application domains as medical domain. One perspective of the paper could be to clarify  how it differs from traditional ontology-based systems and outlining its unique features and advantages in the context of machine learning explainability; for example focusing on the semantic aspects of the provided knowledge base.

\bibliographystyle{IEEEtran}
\bibliography{references}

\end{document}